\documentclass[10pt,twocolumn,letterpaper]{article}

\usepackage{iccv}
\usepackage{times}
\usepackage{epsfig}
\usepackage{graphicx}
\usepackage{amsmath}
\usepackage{amssymb}
\usepackage{comment}
\usepackage{fixltx2e}
\usepackage{caption} 
\usepackage{bm}
\usepackage[dvipsnames]{xcolor}
\usepackage[normalem]{ulem}

\usepackage[colorlinks=true, allcolors=blue]{hyperref}
\usepackage[dvipsnames]{xcolor}

\iccvfinalcopy 

\ificcvfinal\fi

\begin{document}

\title{Attention Based Simple Primitives for Open World Compositional Zero Shot Learning}

\author{Ans Munir\\
Information Technology University\\
Lahore, Pakistan\\
{\tt\small msds20033@itu.edu.pk}
\and
Faisal Z. Qureshi\\
University of Ontario Institute of Technology\\
Oshawa, Canada\\
{\tt\small faisal.qureshi@ontariotechu.ca}
\and
Muhammad Haris Khan\\
MBZUAI\\
Abu Dhabi, UAE\\
{\tt\small muhammad.haris@mbzuai.ac.ae}
\and
Mohsen Ali\\
Information Technology University\\
Lahore, Pakistan\\
{\tt\small mohsen.ali@itu.edu.pk}
}

\maketitle
\ificcvfinal\thispagestyle{empty}\fi

\begin{abstract}
   Compositional Zero-Shot Learning (CZSL) aims to predict  unknown compositions made up of attribute and object pairs. Predicting compositions unseen during training is a challenging  task.
   We are exploring Open World Compositional Zero Shot Learning (OW-CZSL) in this study, where our test space encompasses all potential combinations of attributes and objects. Our approach involves utilizing the self-attention mechanism between attributes and objects to achieve better generalization from seen to unseen compositions. Utilizing a self-attention mechanism facilitates the model's ability to identify relationships between attribute and objects. The similarity between the self-attended textual and visual features is subsequently calculated to generate predictions during the inference phase. The potential test space may encompass implausible object-attribute combinations arising from unrestricted attribute-object pairings. To mitigate this issue, we leverage external knowledge from ConceptNet to restrict the test space to realistic compositions. 
   Our proposed model, Attention-based Simple Primitives (ASP), demonstrates competitive performance, achieving results comparable to the state-of-the-art. Code is available at \textcolor{purple}{https://github.com/ans92/ASP}
 
\end{abstract}

\section{Introduction}

\begin{figure}[t]
\begin{center}
   \includegraphics[width=1.0\linewidth]{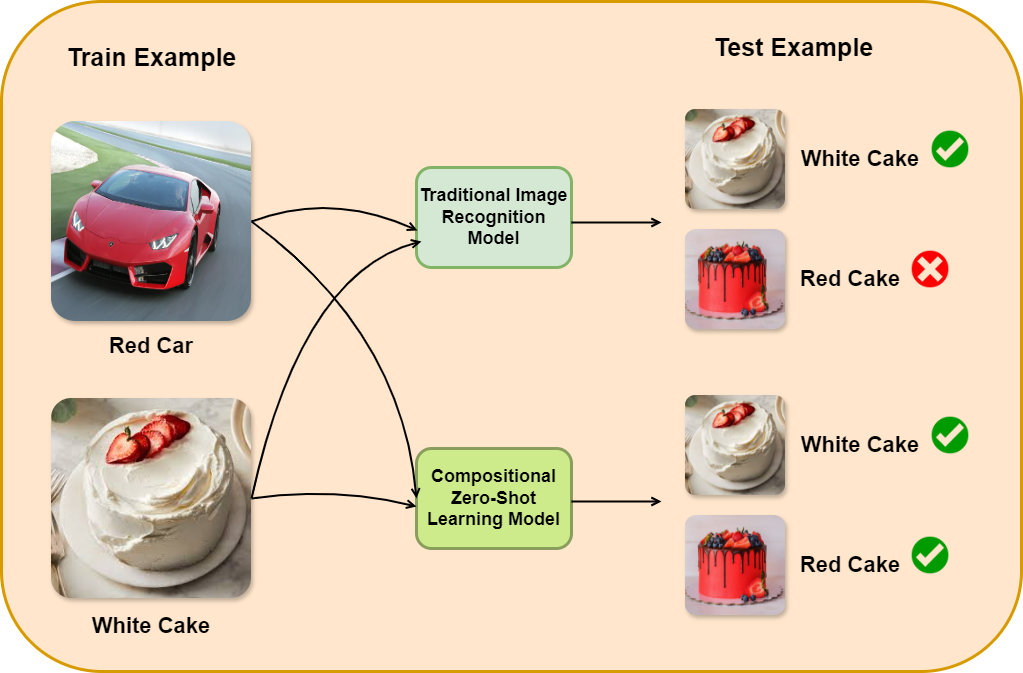}
\end{center}
   \caption{In Compositional Zero-Shot Learning we have training set in the form of compositions that consist of attributes like “Red, White” and objects like “Car, Cake”. Traditional image recognition models typically can only predict known compositions, but they struggle to compose new compositions during testing. In contrast, compositional zero-shot learning model effectively composes new compositions during testing, as evidenced by the example of the ``Red Cake''.}
\label{fig:concept-diagram}
\end{figure}

Humans are capable of recognizing basic concepts in intricate, novel contexts. So, for example, a folded chair looks very different and works very differently from a folded paper, but humans do not have any problem in identifying both with the concept of \emph{folded}. The traditional machine learning and deep learning approaches, however, fail to learn these semantics from the statistical associations that they are typically programmed for. 
The ability to reason about shared and discriminative properties of objects, and identify their attributes is a distinctively human ability, that is very ambitiously mimicked with the task of Compositional Zero Shot Learning (CZSL). 

CZSL aims to learn visual concepts as compositions of attribute and object primitives. For example, a composition \emph{Red Car} consists of attribute \emph{Red} and object \emph{Car}. The formulation of the task of visual learning as compositions of primitives is a paradigm that mimics cognition and finds its applications in a variety of computer vision applications, including but not limited to image retrieval \cite{anwaar2021compositional, neculai2022probabilistic}, visual question-answering \cite{saqur2020multimodal, jing2022maintaining}, and image captioning \cite{nikolaus2019compositional, pantazopoulos2022combine}. This compositionality and contextuality lays the foundation for learning of the long tail of visual concepts very efficiently with a small set of basic primitives.

\begin{figure}[t]
\begin{center}
   \includegraphics[width=1.0\linewidth]{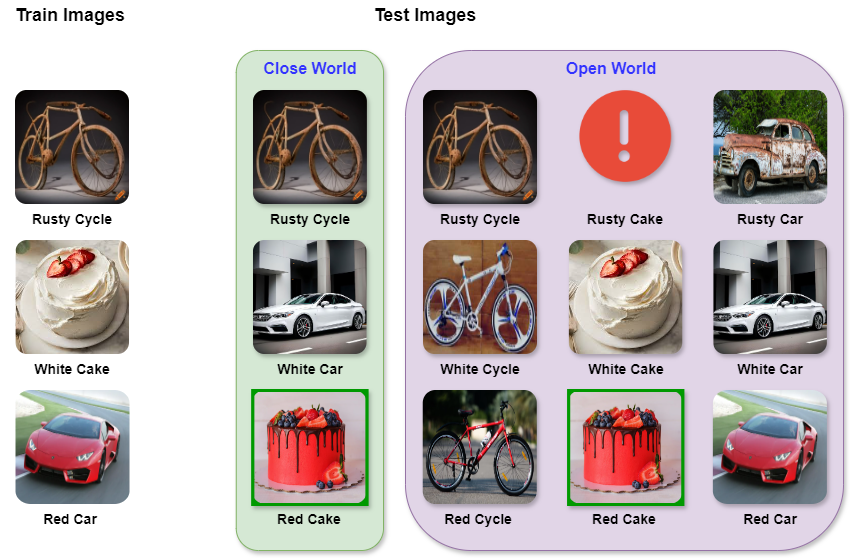}
\end{center}
   \caption{Example demonstrating the difference between close world and open world setting.}
\label{fig:close-world-open-world}

\end{figure}

The CZSL task has two settings: Close World and Open World. 
Traditionally, close world setting has been used to evaluate the models. However, \cite{mancini2021learning} proposed the open world setting.
In a close world setting, its assumed that test-time compositions, are known a-priori. In case of \textit{generalized CZSL}, test-time compositions contain both seen or unseen during training.
A close world situation is depicted in the Figure \ref{fig:close-world-open-world}. Training-set consists of three compositions (\textit{Rusty Cycle, White Cake $\&$ Red Car}) and their respective images. 
However, during the testing-time two more compositions, \textit{White Car} and \textit{Red Cake}, and their images might be present. Therefore, CZSL model's task is to predict the ``Red Cake'' using the five compositions provided in both the training and testing sets.

In  an open world context, the test-time, composition set $C^t$ consists  of all possible combinations of \textit{attributes} and \textit{objects}. 
The $C^t$ is not only quite large, as it grows proportional unique attributes and objects, but might contain compositions that might be \textit{unfeasible} in real-world. \emph{Rusty Cake} is an  example of unfeasible composition, in the  case depicted in Fig. \ref{fig:close-world-open-world}.

\begin{figure}[t]
\begin{center}

   \includegraphics[width=1.0\linewidth]{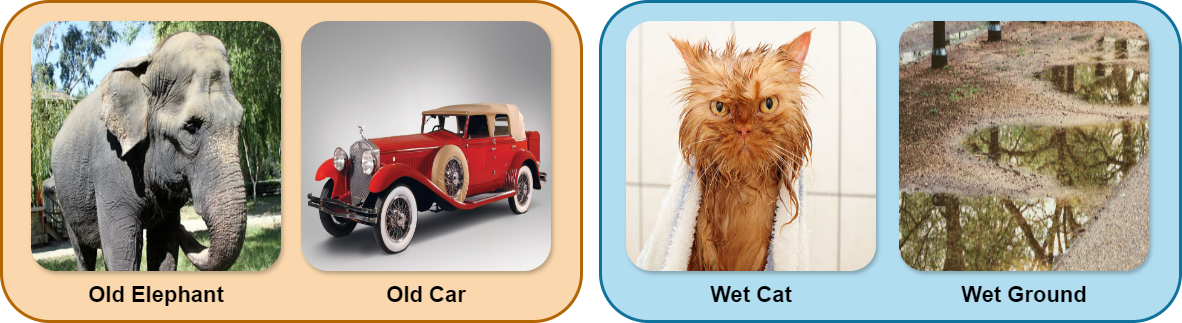}
\end{center}
   \caption[Visual Diversity of Primitive Old and Wet]{Demonstration of Visual Diversity in Primitives. The attribute Old looks drastically different in the context of Elephant (animate object) vs Car (inanimate object). Similarly the attribute Wet looks drastically different in the context of Cat (animate object) vs Ground (inanimate object)}
\label{fig:diversity_old}
\label{fig:onecol}
\end{figure}

CZSL demands two main capacities in a model: (1) the ability to compose and (2) the ability to contextualize. The concept of compositionality has been duly explored in the literature, with a composition of classifiers \cite{misra2017red}, modular networks \cite{purushwalkam2019task}, and composition of text embeddings with the attribute as operator \cite{nagarajan2018attributes} and graph neural networks \cite{naeem2021learning}. But equally important is the ability to contextualize the attributes and objects to new compositions. 
Since these concepts are more semantic than visual, one object can appear drastically different under the influence of different attributes. 
Similarly, the same attribute can appear drastically different in the context of different objects. 
For example, the concept of old manifests with different visual features in the case of an elephant and totally unique visual features in the case of a car, see Figure \ref{fig:diversity_old}. 
Similarly, wet looks different in the context of the ground as compared to that of a cat (Figure \ref{fig:diversity_old}). Therefore, in addition to compositionality, we build a model that captures this contextuality between attributes and objects.

Our main contributions to this paper are:
\begin{itemize}
    
    \item We propose \textbf{A}ttention based \textbf{S}imple \textbf{P}remitives (ASP), a model for open world compositional zero shot learning that uses the self-attention mechanism between attributes and objects to learn interaction between them.
    \item On three benchmark datasets—Mit-States, CGQA, and UT-Zappos—we demonstrated that our model either outperforms or is competitive with prior SOTA results.
\end{itemize}

\section{Related Work}

\noindent\textbf{Compositional zero-shot learning:}
Recently, there has been significant attention given to Compositional Zero Shot Learning (CZSL), which involves predicting joint labels for objects and attributes. Humans demonstrate a remarkable tendency to recognize concepts in different contexts and generalize them to novel contexts. Machines have yet to master this ability.

The earliest approaches in CZSL tried to learn independent classifiers for attribute and object primitives but it failed to capture the joint context of the attribute-object compositions. \cite{misra2017red} learns a transformation network to capture the relation between primitive concepts. 

Many previous works build on the compatibility learning framework, with two major approaches. The first category builds on the idea that attributes lack independent, visual representation and thus treats them as operators on objects. It include works that treat attributes as a linear transformation of objects for learning a composite representation \cite{nagarajan2018attributes}. \cite{li2020symmetry} treats attributes as modifiers of objects while adhering to group axioms, particularly symmetry of object representation under state transformations. 

The second category where the aim is to learn the compatibility score of image embedding with composition embedding spawns a diverse array of approaches. \cite{purushwalkam2019task} learns a modular representation of image embedding, with a gating mechanism conditioned on compositions, which allows modeling of the joint context of the image, object, and state. \cite{wang2019task} develops a conditional generative approach for learning compositional embedding in an adversarial framework. \cite{wei2019adversarial} pools image features from various levels of CNN, and learns compatibility with quintuplet loss under an adversarial framework. CGE \cite{naeem2021learning} learns globally consistent composition embedding, with graph regularization. Compcos \cite{mancini2021open} proposes an open world scenario, where test time predictions are not limited to the set of predefined compositions. Co-CGE \cite{mancini2021learning} combines both the methods, \cite{naeem2021learning} and \cite{mancini2021open} and extends CGE \cite{naeem2021learning} to open world setting. Compcos \cite{mancini2021open} and Co-CGE \cite{mancini2021learning} utilize training compositions to establish the feasibility of compositions in an open world environment.\cite{atzmon2020causal} attempts to uncorrelated the object and attribute representations and deal it with a causal perspective for better generalization.
\cite{ruis2021protoprop} proposes to filter out spurious correlations between attribute and object primitives with independence loss and introduce the desired context in the composition embedding by following it with prototype graph propagation.
\cite{chen2019beyond} uses recurrent attractor networks in the linguistic and visual pathways to improve stability and generalization of composition prototypes, by exploiting the property of convergence and basin of attraction of attractor networks.
\cite{chen2020learning} defines a linguistic graph encoder for information sharing between attributes and objects, for smooth generalization to novel compositions.
\cite{xu2021zero} introduces episodic training for zero-shot framework, and uses cross attention to better capture the joint context of compositions and images. From a contextual perspective, \cite{zoya_thesis} addresses the CZSL issue by conditioning the attributes on objects and vice versa in order to capture the contextual nature of the composition.

Recent work \cite{karthik2022kg} treats attributes and objects separately with independent classifiers to recognize attributes and objects in an open world setting. ConceptNet \cite{conceptnet} has been utilised by them to calculate the feasibility of compositions.
 This kind of independent classifier strategy, as previously mentioned in publications \cite{karthik2022kg}, performs better in an open world context than in a closed one since there are less attributes and objects in an open world context than there are combinations of those attributes and things. 

\noindent\textbf{Attribute classification and zero-shot learning:}
 Very closely tied to the problem of CZSL is the problem of Attribute Classification.
Attribute classification follows the route of learning to describe objects in terms of their attributes, rather than predicting their class labels. It finds use in image retrieval \cite{pham2021learning} and attribute-based zero-shot learning (ZSL) of objects. 
Prototype Learning \cite{xu2020attribute, zhang2017learning, akata2013label} and Generative Models \cite{zhu2018generative, han2020learning, xian2018feature} are two prevalent approaches in ZSL. Score calibration \cite{jayaraman2014decorrelating, le2019classical, liu2018generalized} is also explored for better tuning to unseen classes in ZSL. 

\noindent\textbf{Attention mechanism:}
Self-attention is a building block of Multi-head attention that has been used in transformer architecture and was introduced in \cite{vaswani2017attention}. Attention was originally proposed for Language related models. However, recent works \cite{carion2020end}, \cite{chen2020uniter}, \cite{parmar2018image}, \cite{dosovitskiy2020image} uses attention mechanism for computer vision related tasks. \cite{khan2023learning} uses an attention mechanism for Compositional Zero-Shot Learning by computing attention between compositions. They train and test their model in a closed-world setting.

\noindent\textbf{Our approach} lies at the intersection of different approaches. Like \cite{karthik2022kg} we have predicted attributes and objects independently but in addition to \cite{karthik2022kg} we have also included semantic knowledge \textcolor{blue}{of composition} in our model by allowing attention between attributes and objects. \cite{khan2023learning} also included attention mechanism but they only deal with close world settings. However, unlike \cite{khan2023learning} that has computed attention between compositions, we computed attention between attributes (states) and objects. As we are adding attention between attributes and objects in an open world setting, this makes our approach different from the previous. Different from \cite{zoya_thesis}, our approach involves calculating similarities between attributes and objects using self-attention instead of computing conditional attributes and objects. This will enable us to capitalise on the interaction between attributes and objects.


\begin{figure*}[]
\begin{center}
   \includegraphics[width=1.0\linewidth]{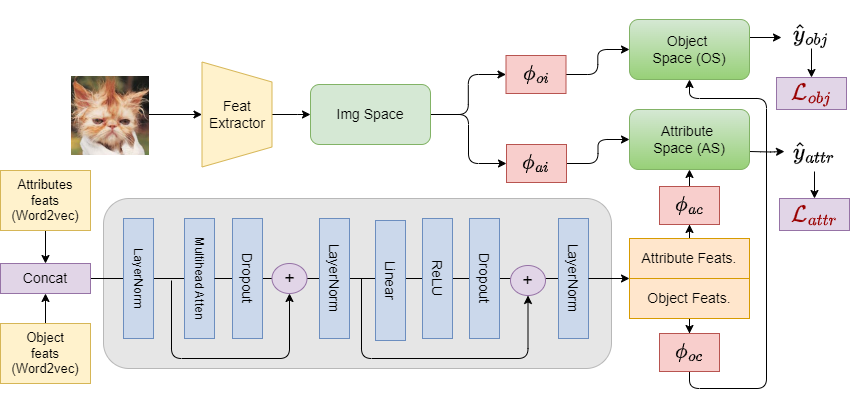}
\end{center}
   \caption{Overall architecture of our ASP model. After concatenating attribute and object features, we compute self-attention between attributes and objects to obtain the interactions between them. Then we get attribute as well as object features and project them to Attribute Space (AS) and Object Space (OS) respectively through respective MLPs. Next, we compute the cosine similarity between attribute image features and composition attribute features. Similarly, we obtain cosine similarity between object image features and composition object features.}
\label{fig:model}

\end{figure*}

\section{Approach}

Proposed CZSL model aims to understand the relationship between attributes, sometimes referred to as states, and objects using the training data.  Subsequently, at test time, the model infers the attribute and object present in an image.  What makes this problem challenging is the fact that the model may not have encountered this attribute/object pair during training time.  Consequently, the model needs to be able to generalize to new attribute/object pairs, which requires the model to capture how attributes and objects interact with each other in a particular visual context.  It is important to note that training data includes all attributes and objects under consideration; however, it often does not contain all possible attributed/object pairs.  See Figure \ref{fig:model} for the overall layout of our model.

\subsection{Problem Formulation}

Given the set of attributes $A = \{ a_1, a_2, ..., a_n \}$ and objects $O = \{ o_1, o_2, ..., o_m \}$, our composition set consists of $C = A \times O$. We deal with the task of visual classification where each image $\bm{x}$  has a compositional label $c$. Specifically in our case the compositional label is a pair of an object primitive $o$ and an attribute primitive $a$, i.e. $c = \{(a,o) | a \in A , o \in O \}$. 

$C_{train}$ is a subset of composition set $C$. The test set $C_{test}$ is defined in three alternate ways in the CZSL literature. In the classic CZSL setting, $C_{test}$ consists of all novel compositions, $C_{train}  \cap  C_{test} = \emptyset$. \cite{purushwalkam2019task} proposed Generalized CZSL setting where test compositions consists of both $C_{train}$ and $C_{test}$ such that $C_{train} \cap C_{test} \neq \emptyset$. Third is an open world setting \cite{naeem2021learning} where test set consists of all the composition set, i.e. $C_{test} = C$

\subsection{Attention Based Simple Primitives}

A feature extractor $w: X \mapsto Z$ computes image features $\bm{z}$ for a given image $\bm{x}$, i.e., $\bm{z} = w( \bm{x} )$. Next, an image-object encoder $\phi_{oi}:Z \mapsto O$ projects image feature $\bm{z}$ to $\bm{o_i}$ within the object space $O$.  Similarly, an image-attribute encoder $\phi_{ai}:Z \mapsto A$ projects image feature $\bm{z}$ to $\bm{a_i}$ within the attribute space $A$.

Word embeddings are used to map object labels and attribute labels into a common space.  The key hypothesis of this work is that the relationship between objects and attributes is not independent of the visual appearance.  We employ multi-headed attention block to capture the relationship between object labels and attributes.  Multi-headed attention processes the object and attribute embeddings and construct transformed object and attribute embeddings.  A object-context encoder $\phi_{oc}$ projects the transformed object label feature to $\bm{c_i}$ within the object space $O$.  Similarly, an attribute-context encoder $\phi{ac}:Z$ projects the transformed attributed label feature to $\bm{a_i}$ within the attribute space $A$. 

Cosine similarity between image object features and object context features is used to predict the object.  Similarly, cosine similarity between image attribute and attribute context is computed to predict the attribute.

The proposed method predicts object labels and attribute labels independently; however, objects and attributes inform each other through the multi-headed attention mechanism mentioned above.

\subsection{Knowledge Guidance for computing feasibility}
To obtain the \textit{composition score} probability of separate predictions of attribute and objects are multiplied together \cite{karthik2022kg}.
As an example, we multiply the scores of ``Red'' and ``Apple'' to obtain the composition's score of ``Red Apple''. We also have a large number of unfeasible compositions.

Not  all of the compositions obtained by combining attributes and objects are feasible. Following  \cite{karthik2022kg}  we employ ConceptNet \cite{conceptnet} to remove unfeasible compositions.
ConceptNet is a type of knowledge graph in which words and phrases are connected by labelled edges.
Therefore, to determine a composition's score, our model compare the score of each attribute to the corresponding object as follows:
\begin{equation}
    c^o_s = \rho_{KB} (s,o)
\end{equation}
where $\rho_{KB} (s,o)$ denotes the score between s and o.

\subsection{Inference}

During inference, once the image features are extracted, we use two multilayer perceptrons (MLPs), $\phi_{si}$ and $\phi_{oi}$, to project image features into two distinct spaces: attribute space and object space.
Similarly, the model project attribute and objects into textual embeddings and then pass them through the attention block to establish attention between attributes and objects. 
Subsequently, our model project these attributes and objects into the attribute space and object space using two MLPs, $\phi_{ac}$ and $\phi_{oc}$, respectively.
In the attribute space, cosine similarity is calculated between attribute visual features and attribute textual features.
Likewise, in the object space, model also calculate the cosine similarity between object visual features and object textual features. The final prediction score is determined by multiplying the attribute and object scores, and the composition with the highest score is predicted as the image label. This process results in the overall prediction function.
\begin{equation}
 f = \underset{(s,o)\in Y}{\arg\max} \Biggl( \phi_{si}(\bm{x}) . \phi_{ac}(s) \Biggr) \times \Biggl( \phi_{oi}(\bm{x}) . \phi_{oc}(o) \Biggr)
\end{equation}
The prediction function mentioned does not use feasibility scores. Feasibility scores are incorporated into the inference function to eliminate unfeasible compositions. After adding the feasibility scores, the inference function becomes,
\begin{equation}
f = \underset{(s,o)\in Y, c^s_o>0}{\arg\max} \Biggl( \phi_{si}(\bm{x}) . \phi_{ac}(s) \Biggr)\times\Biggl( \phi_{oi}(\bm{x}) . \phi_{oc}(o) \Biggr)    
\end{equation}
where $c^s_o>0$ in above function means that we have only included the compositions that have positive feasibility. 

\subsection{Objective Function}
We have used cross-entropy loss for objects as well as attributes to train our model.

\begin{equation}
    \mathcal{L} = \sum^N_{i=1} \mathcal{L}_{state}(x_i, s_i) + \mathcal{L}_{obj}(x_i, o_i)
\end{equation}

\begin{equation*}
    = - \sum_{(x,s)\in T} log \frac{\exp^{p(x,s)}}{\sum_{y \in S} \exp^{p(x,y)}} + \sum_{(x,o)\in T} log \frac{\exp^{p(x,o)}}{\sum_{y \in O} \exp^{p(x,y)}}
\end{equation*} 
where $T$ represents the test set, while $s$ and $o$ are part of the state $S$ and object $O$ sets, respectively.

\begin{table*}[t]\centering
  \begin{tabular}{c|c c c c|c c c c|c c c c}
   &  & MIT-States & & &  & UT-Zappos & & &  & CGQA & & \\
    Method & S & U & HM & AUC & S & U & HM & AUC & S & U & HM & AUC \\ \hline
    
    SymNet \cite{li2020symmetry} & 21.4 & 7.0 & 5.8 & 0.8 & 53.3 & 44.6 & 34.5 & 18.5 & 26.7 & 2.2 & 3.3 & 0.43 \\
    CGE  \textsubscript{ff} \cite{naeem2021learning} & 29.6 & 4.0 & 4.9 & 0.7 & 58.8 & 46.5 & 38.0 & 21.5 & 28.3 & 1.3 & 2.2 & 0.30 \\
    Compcos \cite{mancini2021open} & 25.4 & 10.0 & 8.9 & 1.6 & 59.3 & 46.8 & 36.9 & 21.3 & 28.4 & 1.8 & 2.8 & 0.39 \\
    Co-CGE\textsubscript{ff} \cite{mancini2021learning} & 26.4 & \underline{10.4} & \underline{10.1} & \underline{2.0} & 60.1 & 44.3 & 38.1 & 21.3 & 28.7 & 1.6 & 2.6 & 0.37 \\
    KG-SP\textsubscript{ff} \cite{karthik2022kg} & 23.4 & 7.0 & 6.7 & 1.0 & 58.0 & 47.2 & 39.1 & 22.9 & 26.6 & 2.1 & 3.4 & 0.44 \\
    \textbf{ASP\textsubscript{ff} (ours)} & 21.6 & 7.2 & 6.7 & 1.0 & 59.0 & 42.8 & 38.5 & 21.6 & 26.8 & 2.1 & 3.4 & 0.41 \\
     \hline
    CGE \cite{naeem2021learning} & \textbf{32.4} & 5.1 & 6.0 & 1.0 & \underline{61.7} & 47.7 & 39.0 & 23.1 & \textbf{32.7} & 1.8 & 2.9 & 0.47 \\ 
    Co-CGE \cite{mancini2021learning} & \underline{30.3} & \textbf{11.2} & \textbf{10.7} & \textbf{2.3} & 61.2 & 45.8 & 40.8 & 23.3 & \underline{32.1} & \underline{3.0} & \underline{4.8} & 0.78 \\
    KG-SP \cite{karthik2022kg} & 28.4 & 7.5 & 7.4 & 1.3 & \textbf{61.8} & \textbf{52.1} & \underline{42.3} & \textbf{26.5} & 31.5 & 2.9 & 4.7 & \underline{0.78} \\
    \textbf{ASP (ours)} & 27.1 & 8.4 & 7.7 & 1.4 & 61.0 & \underline{48.6} & \textbf{43.1} & \underline{25.9} & 31.7 & \textbf{3.2} & \textbf{5.0} & \textbf{0.80} 
  \end{tabular}
  \caption{The table compares the outcomes of prior models to ours, ASP. ff indicates that the model employs a fixed feature extractor for image features. The top results are highlighted in bold letters, while the second-best results are underlined. Together with improving results from KG-SP on MIT-States, we have reached state-of-the-art on CGQA. (Higher results are better). Best results in bold and second best underlined.}
  \label{tab:results}
\end{table*}

\section{Experiments}

In this section,  we present the dataset details, evaluation protocols, and implementation details are included. Experimental results comparing previous state-of-the-art methods with proposed methods and ablation study are also included.

\subsection{Experimental Setup}
\noindent\textbf{Datasets:}
Following previous works \cite{mancini2021open}, \cite{mancini2021learning}, \cite{khan2023learning} we tested our model on 3 datasets, MIT-States \cite{StatesAndTransformations}, UT-Zappos \cite{wei2019adversarial,yu2017semantic} and CGQA \cite{naeem2021learning}. 

MIT-States comprises a total of 63,440 images depicting 115 unique attributes and 245 unique object classes. The splits of the dataset are defined as follows, a training set of 30k images constituting 1262 compositions. A validation set of 10k images and a test set of 13k images with composition space equals 28,175 compositions. 

UT-Zappos is the smallest of the 3 datasets and has only 16 attributes and 12 object categories. The splits of ut-Zappos are as follows, a training set consisting of 23k images, and a validation and test set consisting of 3k images. Test time compositional space has 192 compositions.

CGQA was recently proposed in \cite{naeem2021learning} that has a larger number of attributes and objects as compared to MIT-States. It has 413 attributes and 674 object categories. The splits of CGQA are as follows, a training set consisting of 5592 pairs across 26k images; a validation set consisting of 4k images; and a test set consisting of 5k images. Test time compositional space has 278,362 compositions.

\noindent\textbf{Evaluation protocol:}
In our evaluation, we conducted the experiment in an open world setting \cite{mancini2021open, mancini2021learning, karthik2022kg}, including all possible combinations of attributes and objects to form the test set. After following \cite{karthik2022kg, purushwalkam2019task}, we introduced various scalar bias terms to the seen composition scores to address the bias for seen compositions. A large negative scalar bias results in high unseen accuracy and low seen accuracy, while a large positive scalar bias produces the opposite effect. By adjusting the scalar bias from large negative to large positive, we calculated the top-1 accuracy of both seen and unseen compositions. We then plotted the seen-unseen accuracy curve with the seen accuracy on the x-axis and the unseen accuracy on the y-axis. The area under the curve (AUC), best harmonic mean (HM), and the best seen and unseen accuracy were derived from this seen-unseen curve.

\noindent\textbf{Implementation details:}
To extract image features, we used ResNet18 pre-trained on Imagenet as a feature extractor, similar to earlier methods. Two 2-layer MLPs have been employed as classifiers for attributes and objects. In the case of CGQA, we have obtained attributes and object embeddings using Word2Vec. In order to improve comparison with earlier SOTA models, we have employed both Word2Vec \cite{word2vec1, word2vec2} and FastText \cite{fasttext} for the two datasets, MIT-States and UT-Zappos. Then we used an Attention block with one multi-head attention layer followed by Dropout \cite{dropout}, Layer Normalization \cite{layernorm}, Linear layer, and ReLU \cite{relu} activation. Finally, two 2-layer MLPs that map attention features to attribute and object space have also been used. After the first layer, we have utilised LayerNorm, ReLU, and Dropout in every MLP. 

\subsection{Quantitative Analysis}
Results of Open World Compositional Zero Shot Learning (OW-CZSL)  are  shown  in Table \ref{tab:results}. Subscript ''\textsubscript{ff}" in Table~\ref{tab:results} indicates that a fixed feature extractor for image features was not fine-tuned throughout training. The top results shown  in \textbf{bold}, while the outcomes that came in second are underlined. 

Results on MIT-States Dataset are shown in Table \ref{tab:results}.
\cite{mancini2021learning} performs better than every other model in terms of Unseen, HM, and AUC scores. CGE \cite{naeem2021learning} achieved the highest accuracy in terms of seen accuracy. Nonetheless, Unseen, HM, and AUC scores show that our model, ASP, performs better than KG-SP. Our model achieved particularly good results on the MIT-States dataset as compared to KG-SP. Compared to KG-SP's 7.5 unseen accuracy, ASP obtained 8.4. 
Even though the MIT-States dataset is thought to be noisy \cite{atzmon2020causal}, our model performs better than KG-SP.
Table \ref{tab:mit-images} shows the results of our model on MIT-States dataset. First row of Table \ref{tab:mit-images} shows the success examples while second row shows the failure examples of the model. In second row, black labels are actual labels while the red labels are wrong prediction made by our model. 

The most recent dataset for the CZSL challenge is CGQA proposed in \cite{naeem2021learning}. Compared to the other two datasets, this one is the largest and contains the most attributes and objects. In comparison with KG-SP and all other models, our model has attained the highest results. We surpassed all the models including KG-SP in terms of HM, AUC and unseen accuracy. We also obtained a higher seen accuracy compared to KG-SP. Our improved unseen accuracy validates that the attention mechanism of ASP allows for better generalisation to unseen compositions.

In open world environments, the method of independently predicting primitives offers significant advantages. Consider CGQA, the largest dataset, where model only have to predict an image's attribute from 413 attributes and its object from 674 objects. If our model were to predict composition, it would need to select from 278,362 possible combinations, resulting in decreased accuracy as indicated in Table \ref{tab:results} CGQA results. While KG-SP \cite{karthik2022kg} and ASP predict attributes and objects separately, Compcos \cite{mancini2021learning} focuses on predicting compositions. Compared to the other two models, Compcos achieves a lower score.

Although it is be beneficial to independently predict basic elements in an open world setting, the concept of composition remains influential. As demonstrated in Figure \ref{fig:diversity_old}, primitives are interdependent. Nevertheless, we argue that as the dataset expands and encompasses more features and objects, numerous improbable combinations emerge, leading to a decline in model accuracy, as evidenced in the CGQA scenario. As indicated in the results for Compcos in Table \ref{tab:results}, predictive accuracy could be higher when working with a relatively small dataset like MIT-States, but it decreases as the dataset size grows, as observed in the case of CGQA.

The UT-Zappos dataset is small and contains various styles of shoes. Because many types of shoes, such as leather, patent leather, and faux leather, are quite similar, our attention mechanism did not perform well on this specific dataset. Nevertheless, our model still surpasses all other models in the harmonic mean.


\subsection{Qualitative Analysis} 

\begin{table}[t]
\begin{center}
\begin{tabular}{ccc}

\includegraphics[width=0.28\linewidth, height=2.2cm]{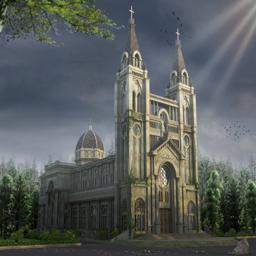} & \includegraphics[width=0.28\linewidth, height=2.2cm]{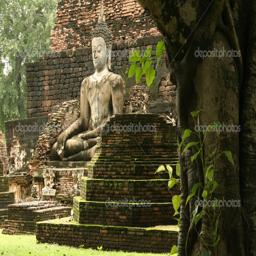} & \includegraphics[width=0.28\linewidth, height=2.2cm]{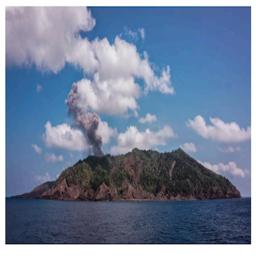} \\
Ancient Church & Ancient Jungle & Barren Island \\

\noalign{\vskip 2mm}  
\includegraphics[width=0.28\linewidth, height=2.2cm]{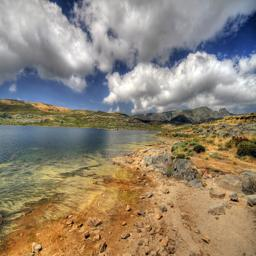} & \includegraphics[width=0.28\linewidth, height=2.2cm]{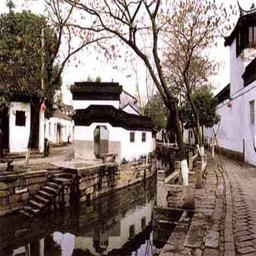} & \includegraphics[width=0.28\linewidth, height=2.2cm]{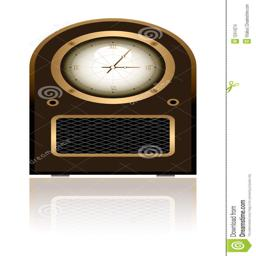} \\

Barren Lake & Ancient Town & Ancient Clock \\

\textcolor{red}{Steaming Lake} & \textcolor{red}{Large House} & \textcolor{red}{Small Clock}

\end{tabular}
\end{center}
\caption{Success examples in first row and failure examples in second row of MIT-States dataset. The ground truth labels are represented by the color black, whereas the model's predictions are indicated by the color red.}
\label{tab:mit-images}
\end{table}


\begin{table}[t]
\begin{center}
\begin{tabular}{ccc}

\includegraphics[width=0.28\linewidth, height=2.2cm]{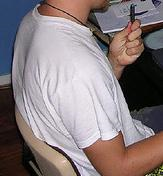} & \includegraphics[width=0.28\linewidth, height=2.2cm]{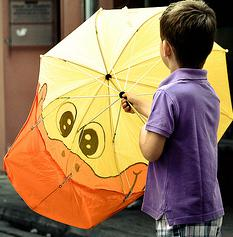} & \includegraphics[width=0.28\linewidth, height=2.2cm]{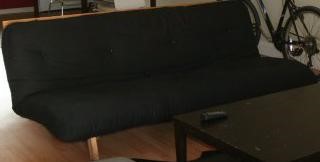} \\
White Shirt & Yellow Umbrella & Black Couch \\

\noalign{\vskip 2mm}  
\includegraphics[width=0.28\linewidth, height=2.2cm]{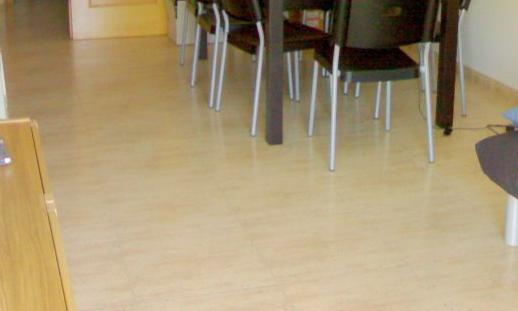} & \includegraphics[width=0.28\linewidth, height=2.2cm]{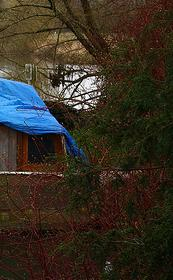} & \includegraphics[width=0.28\linewidth, height=2.2cm]{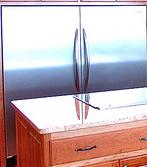} \\

Tan Floor & Red Bush & Large Refrigerator \\

\textcolor{red}{Hardwood Floor} & \textcolor{red}{Large Tree} & \textcolor{red}{Red Window}

\end{tabular}
\end{center}
\caption{Success examples in first row and failure examples in second row of CGQA dataset. The ground truth labels are represented by the color black, whereas the model's predictions are indicated by the color red.}
\label{tab:cgqa-images}
\end{table}

Within this section, we will be showcasing qualitative findings. The initial row in Table \ref{tab:mit-images} depicts instances where our model succeeded, while the second row illustrates instances where it failed. As the images display, the model successfully predicted ``Ancient Church'', ``Ancient Jungle'', and ``Barren Island''. Nonetheless, it inaccurately predicted ``Barren Lake'' as ``Steaming Lake'', ``Ancient Town'' as ``Large House'', and ``Ancient Clock'' as ``Small Clock''. Despite these inaccurate predictions, the model's estimations are still in close proximity to the actual values. In three examples, two have accurately predicted objects, and the predictions are close to ground truth.

In Table \ref{tab:cgqa-images} of the CGQA dataset, the first row showcases our model's successful predictions, including ``White Shirt'', ``Yellow Umbrella'', and ``Black Couch''. The second row, however, displays instances where the model's predictions were not successful. The model correctly identified the object in the first example. In the second example, our model misidentified ``Bush" as a ``Tree'' which are closely related. In the third example, the model's prediction of ``Refrigerator'' as a ``Window'' seems reasonable upon observing the image, as there is a table and the ``Refrigerator'' resembles a window. We can conclude that attention plays a vital role in enabling the model to make more accurate predictions based on the given image.

\subsection{Ablation Study}
\noindent\textbf{Number of heads in Multihead attention:}
Multi-head attention includes many heads. A self-attention system exists inside every head, as demonstrated in \cite{vaswani2017attention}. We have conducted multi-head attention trials with varying head counts. Our studies were conducted on MIT-States using a fixed feature extractor. Additionally, we seeded the values to ensure that the MLPs were initialised uniformly. Our model was trained using 80 epochs for each experiment, and it was then evaluated using the test set.

\begin{figure}[t]
\begin{center}

   \includegraphics[width=1.0\linewidth]{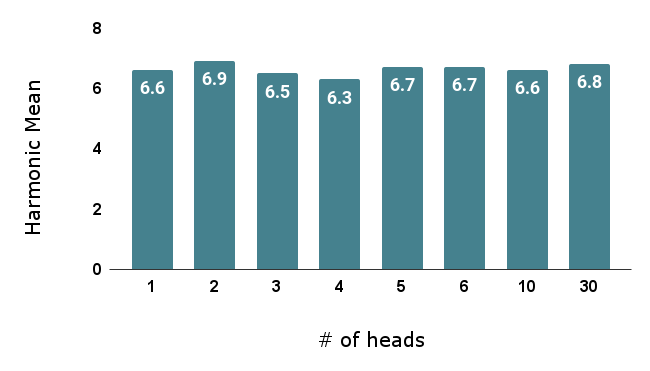}
\end{center}
   \caption{Graph showing the effect of heads in Multihead Attention. On x-axis, there is the number of heads while on the y-axis there is the Harmonic mean.}
\label{fig:multihead-ablation}

\end{figure}

As shown in Fig \ref{fig:multihead-ablation} HM is almost the same in each experiment. The highest accuracy was achieved with two heads in the multi-head attention layer. HM is 6.9 when there are two heads, and 6.8 when there are thirty heads. As a result, accuracy does not increase as the number of heads in the multi-head attention layer grows.

\begin{figure}[t]
\begin{center}
   \includegraphics[width=1.0\linewidth]{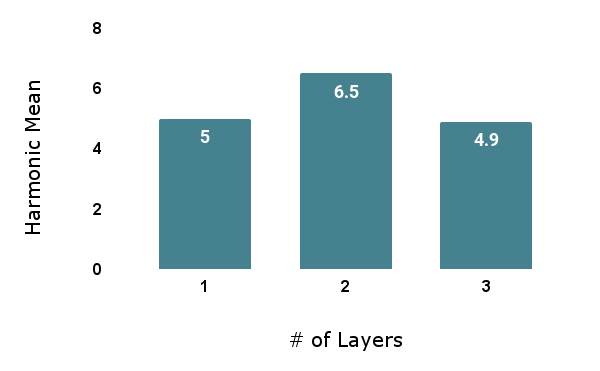}
\end{center}
   \caption{Graph showing the effect of the number of linear layers in all the MLPs on the Harmonic Mean.}
\label{fig:layers-hm}

\end{figure}

\noindent\textbf{Depth of MLPs:}
We have examined how each MLP's depth affects MIT-States using fixed feature extractors. The impact of the number of linear layers in an MLP on the harmonic mean is depicted in Fig. \ref{fig:layers-hm}. It is evident that a single layer prevents the model from understanding the intricate relationships between the attributes and objects. Nevertheless, the model is able to effectively learn the intricate relationships between them when the number of layers is increased to two. However, as we increased the number of layers, the model model likely overfits the data and hence its Harmonic Mean decreased.

\section{Conclusion}

We propose a  novel  solution for  the  Compositional Zero Shot Learning (CZSL) in the Open-World setting. Unlike in closed-world CZSL, where composition set during  test is known in advance, open-world setting assume no  prior  knowledge therefore include all possible combinations of attributes and objects. 
Primitives, such as attributes and objects, have been predicted independently. 
Our model has employed attention mechanism between primitives to capture their interactions, to generalize effectively from training compositions to test compositions.
Our approach of independently predicting primitives performs better due to the significant reduction in our test space in open-world CZSL. 
In the open-world CZSL, there are also unrealistic compositions which we have addressed by utilizing external knowledge from ConceptNet. 
Our model achieved the top performance on the CGQA dataset, surpassing all previous models and achieving higher accuracy than KG-SP.


{\small
\bibliographystyle{ieee_fullname}
\bibliography{egbib}

\begin{thebibliography}{10}\itemsep=-1pt

\bibitem{akata2013label}
Zeynep Akata, Florent Perronnin, Zaid Harchaoui, and Cordelia Schmid.
\newblock Label-embedding for attribute-based classification.
\newblock In {\em Proceedings of the IEEE conference on computer vision and pattern recognition}, pages 819--826, 2013.

\bibitem{anwaar2021compositional}
Muhammad~Umer Anwaar, Egor Labintcev, and Martin Kleinsteuber.
\newblock Compositional learning of image-text query for image retrieval.
\newblock In {\em Proceedings of the IEEE/CVF Winter conference on Applications of Computer Vision}, pages 1140--1149, 2021.

\bibitem{atzmon2020causal}
Yuval Atzmon, Felix Kreuk, Uri Shalit, and Gal Chechik.
\newblock A causal view of compositional zero-shot recognition.
\newblock In {\em Advances in Neural Information Processing Systems (NeurIPS)}, 2020.

\bibitem{layernorm}
Jimmy~Lei Ba, Jamie~Ryan Kiros, and Geoffrey~E Hinton.
\newblock Layer normalization.
\newblock {\em arXiv preprint arXiv:1607.06450}, 2016.

\bibitem{fasttext}
Piotr Bojanowski, Edouard Grave, Armand Joulin, and Tomas Mikolov.
\newblock Enriching word vectors with subword information.
\newblock {\em Transactions of the association for computational linguistics}, 5:135--146, 2017.

\bibitem{carion2020end}
Nicolas Carion, Francisco Massa, Gabriel Synnaeve, Nicolas Usunier, Alexander Kirillov, and Sergey Zagoruyko.
\newblock End-to-end object detection with transformers.
\newblock In {\em Computer Vision--ECCV 2020: 16th European Conference, Glasgow, UK, August 23--28, 2020, Proceedings, Part I 16}, pages 213--229. Springer, 2020.

\bibitem{chen2020learning}
Hui Chen, Zhixiong Nan, Jingjing Jiang, and Nanning Zheng.
\newblock Learning to infer unseen attribute-object compositions.
\newblock {\em arXiv preprint arXiv:2010.14343}, 2020.

\bibitem{chen2019beyond}
Hui Chen, Zhixiong Nan, and Nanning Zheng.
\newblock Beyond supervised learning: Recognizing unseen attribute-object pairs with vision-language fusion and attractor networks.
\newblock 2019.

\bibitem{chen2020uniter}
Yen-Chun Chen, Linjie Li, Licheng Yu, Ahmed El~Kholy, Faisal Ahmed, Zhe Gan, Yu Cheng, and Jingjing Liu.
\newblock Uniter: Universal image-text representation learning.
\newblock In {\em Computer Vision--ECCV 2020: 16th European Conference, Glasgow, UK, August 23--28, 2020, Proceedings, Part XXX}, pages 104--120. Springer, 2020.

\bibitem{dosovitskiy2020image}
Alexey Dosovitskiy, Lucas Beyer, Alexander Kolesnikov, Dirk Weissenborn, Xiaohua Zhai, Thomas Unterthiner, Mostafa Dehghani, Matthias Minderer, Georg Heigold, Sylvain Gelly, et~al.
\newblock An image is worth 16x16 words: Transformers for image recognition at scale.
\newblock {\em arXiv preprint arXiv:2010.11929}, 2020.

\bibitem{han2020learning}
Zongyan Han, Zhenyong Fu, and Jian Yang.
\newblock Learning the redundancy-free features for generalized zero-shot object recognition.
\newblock In {\em Proceedings of the IEEE/CVF Conference on Computer Vision and Pattern Recognition}, pages 12865--12874, 2020.

\bibitem{zoya_thesis}
Zoya~Naseer Hashmi.
\newblock Conditional embeddings for compositional zero shot learning.
\newblock Master's thesis, Information Technology University, Lahore, 2022.

\bibitem{StatesAndTransformations}
Phillip Isola, Joseph~J. Lim, and Edward~H. Adelson.
\newblock Discovering states and transformations in image collections.
\newblock In {\em CVPR}, 2015.

\bibitem{jayaraman2014decorrelating}
Dinesh Jayaraman, Fei Sha, and Kristen Grauman.
\newblock Decorrelating semantic visual attributes by resisting the urge to share.
\newblock In {\em Proceedings of the IEEE Conference on Computer Vision and Pattern Recognition}, pages 1629--1636, 2014.

\bibitem{jing2022maintaining}
Chenchen Jing, Yunde Jia, Yuwei Wu, Xinyu Liu, and Qi Wu.
\newblock Maintaining reasoning consistency in compositional visual question answering.
\newblock In {\em Proceedings of the IEEE/CVF Conference on Computer Vision and Pattern Recognition}, pages 5099--5108, 2022.

\bibitem{karthik2022kg}
Shyamgopal Karthik, Massimiliano Mancini, and Zeynep Akata.
\newblock Kg-sp: Knowledge guided simple primitives for open world compositional zero-shot learning.
\newblock In {\em Proceedings of the IEEE/CVF Conference on Computer Vision and Pattern Recognition}, pages 9336--9345, 2022.

\bibitem{khan2023learning}
Muhammad Gul Zain~Ali Khan, Muhammad~Ferjad Naeem, Luc Van~Gool, Alain Pagani, Didier Stricker, and Muhammad~Zeshan Afzal.
\newblock Learning attention propagation for compositional zero-shot learning.
\newblock In {\em Proceedings of the IEEE/CVF Winter Conference on Applications of Computer Vision}, pages 3828--3837, 2023.

\bibitem{le2019classical}
Yannick Le~Cacheux, Herv{\'e} Le~Borgne, and Michel Crucianu.
\newblock From classical to generalized zero-shot learning: A simple adaptation process.
\newblock In {\em International Conference on Multimedia Modeling}, pages 465--477. Springer, 2019.

\bibitem{li2020symmetry}
Yong-Lu Li, Yue Xu, Xiaohan Mao, and Cewu Lu.
\newblock Symmetry and group in attribute-object compositions.
\newblock In {\em Proceedings of the IEEE/CVF Conference on Computer Vision and Pattern Recognition}, pages 11316--11325, 2020.

\bibitem{liu2018generalized}
Shichen Liu, Mingsheng Long, Jianmin Wang, and Michael~I Jordan.
\newblock Generalized zero-shot learning with deep calibration network.
\newblock {\em Advances in Neural Information Processing Systems}, 31, 2018.

\bibitem{mancini2021learning}
Massimiliano Mancini, Muhammad Ferjad~Naeem, Yongqin Xian, and Zeynep Akata.
\newblock Learning graph embeddings for open world compositional zero-shot learning.
\newblock {\em arXiv e-prints}, pages arXiv--2105, 2021.

\bibitem{mancini2021open}
M Mancini, MF Naeem, Y Xian, and Zeynep Akata.
\newblock Open world compositional zero-shot learning.
\newblock In {\em 34th IEEE Conference on Computer Vision and Pattern Recognition}. IEEE, 2021.

\bibitem{word2vec1}
Tomas Mikolov, Kai Chen, Greg Corrado, and Jeffrey Dean.
\newblock Efficient estimation of word representations in vector space.
\newblock {\em arXiv preprint arXiv:1301.3781}, 2013.

\bibitem{word2vec2}
Tomas Mikolov, Ilya Sutskever, Kai Chen, Greg~S Corrado, and Jeff Dean.
\newblock Distributed representations of words and phrases and their compositionality.
\newblock {\em Advances in neural information processing systems}, 26, 2013.

\bibitem{misra2017red}
Ishan Misra, Abhinav Gupta, and Martial Hebert.
\newblock From red wine to red tomato: Composition with context.
\newblock In {\em Proceedings of the IEEE Conference on Computer Vision and Pattern Recognition}, pages 1792--1801, 2017.

\bibitem{naeem2021learning}
MF Naeem, Y Xian, F Tombari, and Zeynep Akata.
\newblock Learning graph embeddings for compositional zero-shot learning.
\newblock In {\em 34th IEEE Conference on Computer Vision and Pattern Recognition}. IEEE, 2021.

\bibitem{nagarajan2018attributes}
Tushar Nagarajan and Kristen Grauman.
\newblock Attributes as operators: factorizing unseen attribute-object compositions.
\newblock In {\em Proceedings of the European Conference on Computer Vision (ECCV)}, pages 169--185, 2018.

\bibitem{relu}
Vinod Nair and Geoffrey~E Hinton.
\newblock Rectified linear units improve restricted boltzmann machines.
\newblock In {\em ICML}, 2010.

\bibitem{neculai2022probabilistic}
Andrei Neculai, Yanbei Chen, and Zeynep Akata.
\newblock Probabilistic compositional embeddings for multimodal image retrieval.
\newblock In {\em Proceedings of the IEEE/CVF conference on computer vision and pattern recognition}, pages 4547--4557, 2022.

\bibitem{nikolaus2019compositional}
Mitja Nikolaus, Mostafa Abdou, Matthew Lamm, Rahul Aralikatte, and Desmond Elliott.
\newblock Compositional generalization in image captioning.
\newblock {\em arXiv preprint arXiv:1909.04402}, 2019.

\bibitem{pantazopoulos2022combine}
George Pantazopoulos, Alessandro Suglia, and Arash Eshghi.
\newblock Combine to describe: Evaluating compositional generalization in image captioning.
\newblock In {\em Proceedings of the 60th Annual Meeting of the Association for Computational Linguistics: Student Research Workshop}, pages 115--131, 2022.

\bibitem{parmar2018image}
Niki Parmar, Ashish Vaswani, Jakob Uszkoreit, Lukasz Kaiser, Noam Shazeer, Alexander Ku, and Dustin Tran.
\newblock Image transformer.
\newblock In {\em International conference on machine learning}, pages 4055--4064. PMLR, 2018.

\bibitem{pham2021learning}
Khoi Pham, Kushal Kafle, Zhe Lin, Zhihong Ding, Scott Cohen, Quan Tran, and Abhinav Shrivastava.
\newblock Learning to predict visual attributes in the wild.
\newblock In {\em Proceedings of the IEEE/CVF Conference on Computer Vision and Pattern Recognition}, pages 13018--13028, 2021.

\bibitem{purushwalkam2019task}
Senthil Purushwalkam, Maximilian Nickel, Abhinav Gupta, and Marc'Aurelio Ranzato.
\newblock Task-driven modular networks for zero-shot compositional learning.
\newblock In {\em Proceedings of the IEEE/CVF International Conference on Computer Vision}, pages 3593--3602, 2019.

\bibitem{ruis2021protoprop}
Frank Ruis, Gertjan~J Burghouts, and Doina Bucur.
\newblock Independent prototype propagation for zero-shot compositionality.
\newblock In {\em Advances in Neural Information Processing Systems (NeurIPS)}, volume~34, 2021.

\bibitem{saqur2020multimodal}
Raeid Saqur and Karthik Narasimhan.
\newblock Multimodal graph networks for compositional generalization in visual question answering.
\newblock {\em Advances in Neural Information Processing Systems}, 33:3070--3081, 2020.

\bibitem{conceptnet}
R Speer, J Chin, and C~ConceptNet Havasi.
\newblock 5.5: An open multilingual graph of general knowledge.
\newblock In {\em Proceedings of the Thirty-First AAAI Conference on Artificial Intelligence (December 2016)}, pages 4444--4451.

\bibitem{dropout}
Nitish Srivastava, Geoffrey Hinton, Alex Krizhevsky, Ilya Sutskever, and Ruslan Salakhutdinov.
\newblock Dropout: a simple way to prevent neural networks from overfitting.
\newblock {\em The journal of machine learning research}, 15(1):1929--1958, 2014.

\bibitem{vaswani2017attention}
Ashish Vaswani, Noam Shazeer, Niki Parmar, Jakob Uszkoreit, Llion Jones, Aidan~N Gomez, {\L}ukasz Kaiser, and Illia Polosukhin.
\newblock Attention is all you need.
\newblock {\em Advances in neural information processing systems}, 30, 2017.

\bibitem{wang2019task}
Xin Wang, Fisher Yu, Trevor Darrell, and Joseph~E Gonzalez.
\newblock Task-aware feature generation for zero-shot compositional learning.
\newblock {\em arXiv preprint arXiv:1906.04854}, 2019.

\bibitem{wei2019adversarial}
Kun Wei, Muli Yang, Hao Wang, Cheng Deng, and Xianglong Liu.
\newblock Adversarial fine-grained composition learning for unseen attribute-object recognition.
\newblock In {\em Proceedings of the IEEE/CVF International Conference on Computer Vision}, pages 3741--3749, 2019.

\bibitem{xian2018feature}
Yongqin Xian, Tobias Lorenz, Bernt Schiele, and Zeynep Akata.
\newblock Feature generating networks for zero-shot learning.
\newblock In {\em Proceedings of the IEEE conference on computer vision and pattern recognition}, pages 5542--5551, 2018.

\bibitem{xu2021zero}
Guangyue Xu, Parisa Kordjamshidi, and Joyce~Y Chai.
\newblock Zero-shot compositional concept learning.
\newblock {\em arXiv preprint arXiv:2107.05176}, 2021.

\bibitem{xu2020attribute}
Wenjia Xu, Yongqin Xian, Jiuniu Wang, Bernt Schiele, and Zeynep Akata.
\newblock Attribute prototype network for zero-shot learning.
\newblock {\em Advances in Neural Information Processing Systems}, 33:21969--21980, 2020.

\bibitem{yu2017semantic}
Aron Yu and Kristen Grauman.
\newblock Semantic jitter: Dense supervision for visual comparisons via synthetic images.
\newblock In {\em Proceedings of the IEEE International Conference on Computer Vision}, pages 5570--5579, 2017.

\bibitem{zhang2017learning}
Li Zhang, Tao Xiang, and Shaogang Gong.
\newblock Learning a deep embedding model for zero-shot learning.
\newblock In {\em Proceedings of the IEEE conference on computer vision and pattern recognition}, pages 2021--2030, 2017.

\bibitem{zhu2018generative}
Yizhe Zhu, Mohamed Elhoseiny, Bingchen Liu, Xi Peng, and Ahmed Elgammal.
\newblock A generative adversarial approach for zero-shot learning from noisy texts.
\newblock In {\em Proceedings of the IEEE conference on computer vision and pattern recognition}, pages 1004--1013, 2018.

\end{thebibliography}
}

\end{document}